# Equitable Community Resilience: The Case of Winter Storm Uri in Texas

Ali Nejat[1], Ph.D., P.E., PMP, Associate Professor
*Texas Tech University*

Laura Solitare, Ph.D., Associate Professor
*Texas Southern University*

Edward Pettitt, MPH, Graduate Student
*Texas Southern University*

Hamed Mohsenian-Rad, Ph.D., IEEE Fellow, Bourns Family Fellow, Professor
*University of California, Riverside*

## ABSTRACT

Community resilience in the face of natural hazards relies on a community's potential to bounce back. A failure to integrate equity into resilience considerations results in unequal recovery and disproportionate impacts on vulnerable populations, which has long been a concern in the United States. This research investigated aspects of equity related to community resilience in the aftermath of Winter Storm Uri in Texas which led to extended power outages for more than 4 million households. County-level outage/recovery data was analyzed to explore potential significant links between various county attributes and their share of the outages during the recovery/restoration phase. Next, satellite imagery was used to examine data at a much higher geographical resolution focusing on census tracts in the city of Houston. The goal was to use computer vision to extract the extent of outages within census tracts and investigate their linkages to census tracts attributes. Results from various statistical procedures revealed statistically significant negative associations between counties' percentage of non-Hispanic whites and median household income with the ratio of outages. Additionally, at census tract level, variables including percentages of *linguistically isolated* population and *public transport users* exhibited positive associations with the group of census tracts that were affected by the outage as detected by computer vision analysis. Informed by these results, engineering solutions such as the applicability of grid modernization technologies, together with distributed and renewable energy resources, when controlled for the region's topographical characteristics, are proposed to enhance equitable power grid resiliency in the face of natural hazards.

## INTRODUCTION

Between February 13-17, 2021, Winter Storm Uri impacted 25 states and more than 150 million Americans leading to extended power and water outages nationwide, with Texas being the hardest hit (Centers, 2021). The impact of Uri on the state of Texas was far beyond expectation due to the lack of precautionary measures to withstand prolonged freezing temperatures (Englund, 2021). While the electric grid in Texas is capable of withstanding extreme humidity and warm weather conditions, it was not designed to endure extended freezing temperatures, leaving administrators with no other options than to implement rolling blackouts. These blackouts were supposed to last less than an hour (Philipose, 2021) but ranged anywhere from couple of hours to couple of days across the state (Ferman, Sparber, & Limón, 2021; Miller, 2021). The storm led to several deaths due to carbon monoxide poisoning, a significant halt in the administration of COVID-19 vaccines, and an economic loss that is estimated to be around $90 billion (Sullivan, 2021). Meanwhile, various outlets brought to light the issue of environmental justice, and the human-centric and equity aspects of the impact (Stevens, 2021). While four million people lost electricity and water in Texas (Srikanth, 2021; Stevens, 2021), the impact was disproportionate in low-income communities of color (Curtis & Campbell, 2021; SolarOne, 2021; Yancey-Bragg & Jervis, 2021). Low-income non-white families were reported to bear the brunt of compounding crises in the aftermath of Uri (SolarOne, 2021).

Disparate impact was also investigated by Carvallo et al (2021) who used satellite data on nighttime lights (NL) to determine blackouts at the level of Census Block Groups (CBGs) and correlated it to demographic data from the EPA's EJScreen tool which identified the share of population in each CBG that is minority and low-income; they found that CBGs with a high share of non-white minorities were more than four times as likely to experience a blackout than predominantly white CBGs, regardless of income. Lee et al (2021) analyzed community-scale big data, including digital trace and crowdsourced data, in Harris County and found significant disparity in the extent and duration of power outages as well as the extent of burst pipes and disrupted food access for low-income and minority residents at the census tract level.

---

[1] Corresponding Author, Email: ali.nejat@ttu.edu

Racial and income disparities in Texas are associated with a history of redlining and gentrification (Srikanth, 2021; Stevens, 2021) resulting in the concentration of people of color in specific areas. At the same time, low-income areas tend to lack amenities such as hospitals, which during the blackout were prioritized for power restoration together with their neighboring areas (Stevens, 2021). Additionally, tendency to live in neighborhoods with older homes equipped with utility systems more prone to failure was cited along with other contributing factors to this disproportionate impact for Texans of color (Ura & Garnham, 2021).

A disparity in environmental justice was also witnessed in the aftermath of other recent storms including Sandy (Burger, Gochfeld, & Lacy, 2019; Carbone & Wright, 2016; Lieberman-Cribbin, Gillezeau, Schwartz, & Taioli, 2020), Katrina (Birch & Wachter, 2006; Frey & Singer, 2006; Zoraster, 2010), Harvey (Bodenreider, Wright, Barr, Xu, & Wilson, 2019; Chakraborty, Collins, & Grineski, 2019; Collins, Grineski, Chakraborty, & Flores, 2019) and many others. Given the fact that the disproportionate impact of disasters on vulnerable populations including low-income non-white families is not unique to Uri nor to the state of Texas, there is a dire need to demystify its contributing factors using appropriate data.

Such s study can provide opportunities to learn from what went wrong at various levels, including technical upkeep of infrastructure, precautionary measures, planning, and execution with the hope that it can be used to foster environmental justice for future community resilience planning (Wang, Chen, Wang, & Baldick, 2015; Zhang & Torres, 2020). This research intends to address this issue through collecting and extracting data at various county and census tract levels and contribute to the existing body of knowledge from multiple fronts. More specifically, unlike previous studies, our study not only revealed the importance of household-specific attributes such as education and language but also highlighted the significance of the built environment through parameters such as the percentage of one-unit structures and public-transport users within census tracts. This section is succeeded by Research Methodology, and Results sections. Conclusions will wrap up the paper.

**METHODOLOGY**

To address the objectives of this research, data analysis was performed at both county and census tract levels. County level analysis was made available through the data that was acquired from a third-party company[2] that is specialized in collecting live power outage data from utilities nationwide. The first round of data analysis was performed at county-level to explore changes in power outages and potentially link them to county-specific characteristics. To do so, a macro analysis of longitudinal county-level outages during the month of February was studied. The study was mainly focused on the *recovery stage* of the blackout, which, as shown in Figure 1b, happened between Feb. 18 – Feb. 21. The rationale behind using the *recovery* phase was the implication of human interventions through the decisions of grid operation professionals and restoration plans/policies (Koutsoukis, Georgilakis, & Hatziargyriou, 2019; Liu, Fan, & Terzija, 2016; National Academies of Sciences & Medicine, 2017), which in general contrasts with the *build-up* and *blackout* stages in which decisions could be made on an ad-hoc basis. Additionally, Figure 1a shows total electric load in the ERCOT network (in GW) over a period of three weeks from February 5 until February 26. Abnormal trends in increased loading starts around February 9 due to record low temperature and peaks around February 15 (ERCOT, 2021a); Figure 1b represents the subsequent severe blackout that affects between 2.5 million to 4.5 million customers over a period of three days together with build-up and recovery which starts around February 18. Finally, Figure 1c shows a snapshot map of the power outages on February 20 at noon and the extent of impact (PowerOutage.US, 2021). Full recovery took more days to be achieved at certain locations across the state.

---

[2] Bluefire Studios LLC https://poweroutage.us/about

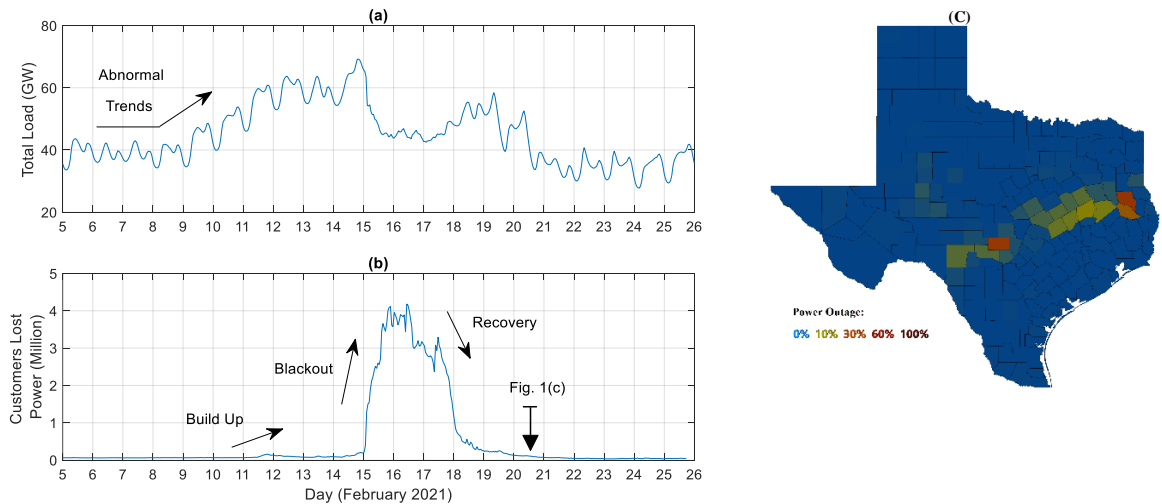

*Figure 1. Overview of the power grid conditions during Winter Storm Uri in Texas:*

Next, to explore the impacts at a neighborhood scale, census tract data for Harris County was used in the second round of data analysis, including the variables shown in Table 1. Our statistical tests used census variables from the 2019 American Communities Survey (ACS 2019) 5-year estimates for Harris County Census Tracts. We selected variables that are proxies for environmental and social disparities at the neighborhood level (we use the term neighborhood interchangeably with census tract). Once data at census tract level was obtained, the next step was to identify a measure that could be used as a proxy for extent of outages within the census tracts. In the absence of public access to high resolution outage data, nightlight satellite imagery[3] was used for this purpose. The idea was to compare the total number of black and white pixels within each census tract before and after the winter storm and evaluate the changes.

*Table 1.* Census Tract Variables Extracted from 2019 ACS

| Social Characteristics | Housing Characteristics | Economic Characteristics |
| --- | --- | --- |
| *Race* | *Residential type* | *Median Household Income* |
| *Ethnicity* | *Vacant housing* | *Unemployment* |
| *Education* | *Owner Occupancy* | *Poverty* |
| *Linguistically Isolated* | *Mobility* | *Employment Industry* |
| | | *Access to Health Insurance* |
| | | *Transportation modes to commute to work* |

**RESULTS**

**Data Analysis at County Level**

At county level, our study was focused on the time span during which about 10% of the total outages (~400K units) had not yet been recovered. The primary objective was to explore potential significant links between different counties' attributes and their share of the *remaining outages* during this *final recovery/restoration phase*. Results are shown in Figure 2 in which linear regression models were developed to investigate the significance of various county-level attributes, including counties' *percentage of non-Hispanic whites* and *median household income* in driving counties' share of the remaining outages. These preliminary results revealed statistically significant negative associations between these predictors and the dependent variable, which was set to be *the ratio of remaining outages to the maximum recorded outage in the county*. This would imply that counties with higher percentage of *non-Hispanic whites* recovered faster (lower rate of remaining outages) compared to those without. The same result can be concluded from counties with higher median household income. These results are aligned with the existing literature on the higher vulnerability of non-white and lower income groups within the context of disasters (Cutter, Boruff, & Shirley, 2003; Grube, Fike, & Storr, 2018; Zahran, Brody, Peacock, Vedlitz, & Grover, 2008). Even though results from this preliminary study were thought provoking, they were not conclusive and, as such, called for more in-depth bottom-up analyses to be performed. We subsequently performed a micro-level analysis of power outages at census tract level within Harris County due to its high level of impact among cities within the state of Texas (HSPA, 2021).

---

[3] https://www.newsweek.com/satellite-photos-show-extent-texas-power-outages-space-1569942

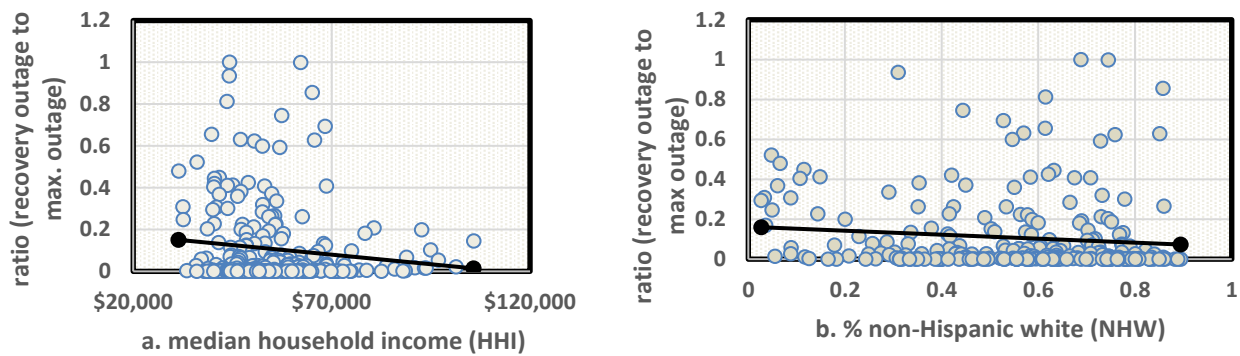

*Figure 2. Preliminary Study Results Based on County-level Outage Data*

**Data analysis at Census Tract Level**

At census tract level, tracts' attributes as shown in Table 1 were extracted from ACS 2019 and were associated with results from computer vision analysis of their extent of outage. More specifically, in the absence of high-resolution longitudinal power outage data within greater Harris County, the following tasks as demonstrated in Figure 3 were performed to use satellite imagery to detect the intensity of outages. First, NASA satellite images (NASA, 2021) of nighttime lights in Texas on Feb. 7 and Feb. 16 2021 were downloaded from their website (Fig. 3a). The image taken on Feb. 7 was set as the benchmark displaying normal nighttime lights while the second image taken on Feb. 16 was used to evaluate changes during the outage. Second, to perform spatial analysis, downloaded images were georeferenced in ArcGIS using ArcMap georeferencing tool (Fig. 3b). Third, census tracts shapefile was downloaded from US Census TIGER/Line[4] and clipped by Harris County boundaries in ArcMap (Fig. 3c). Fourth, ArcMap model builder was used to extract multiple shapefiles for each census tract and create raster files from the extracted shapefiles (Fig. 3d).

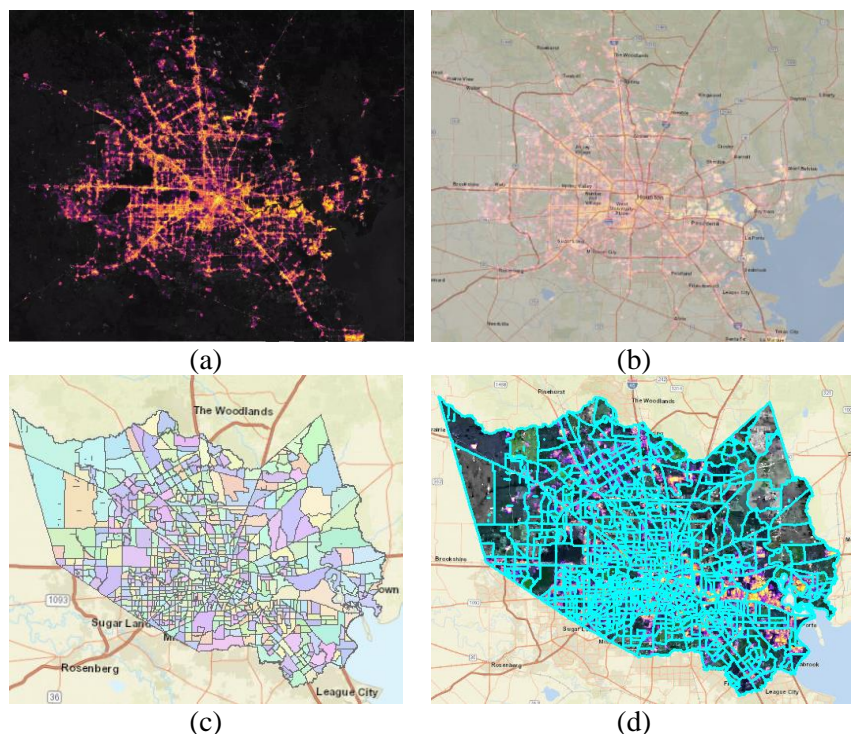

*Figure 3. Census Tract Computer Vision Process*

Once raster datasets were created for both before the outage and after the outage, OpenCV computer vision package[5] was used to convert the datasets from Red-Green-Blue (RGB) to grayscale and then to black and white (see Figure 4) and to count the number of pixels within each census tract. These numbers were the basis for calculating *black pixel ratio* which was used to determine the severity of outage within neighborhoods for the rest of the research study.

---
[4] https://www.census.gov/cgi-bin/geo/shapefiles/index.php?year=2019&layergroup=Census+Tracts
[5] https://opencv.org/releases/

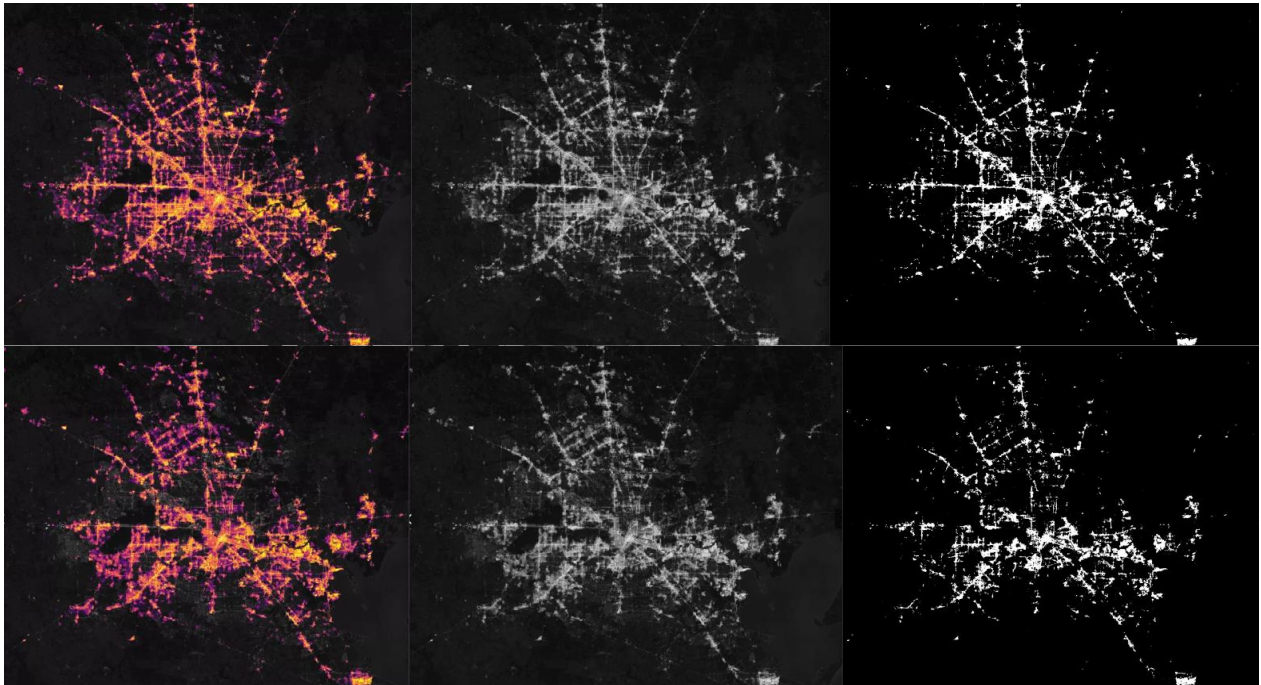

*Figure 4. Preliminary Study Results Based on County-level Outage Data*

*Significant Variables and their Effect Size*

Once outage intensity at census tract level was determined using satellite photos and computer vision algorithms, independent sample t-tests were performed to understand the differences between the two sample groups: neighborhoods (census tracts) that experienced a power outage compared to neighborhoods (census tracts) that did not have a power outage during Winter Storm Uri. As shown in Table 2, these tests were run using various measures to define which neighborhoods had a power outage. A key challenge in this analysis is to lower the sensitivity to normal fluctuation in customer loads, a phenomenon that is well documented in the power engineering literature due to the impact of various factors on load profiles (Mohsenian-Rad, 2021). Importantly, even if there was no blackout at the times when either of the two satellite images were taken, one could still have seen a considerable level of difference between the light pattern and intensity across the two satellite images unrelated to any outage. Of course, such normal fluctuations also exist when a comparison is made between satellite images *before* and *during* outages due to Winter Storm Uri; yet again, unrelated to the outages. Therefore, a reduction in the sensitivity of the analysis to such normal fluctuations in loading was needed.

To resolve this issue, a reliability threshold in the form of a cut-off point to be applied to the black pixel ratio was considered. First, suppose there is no normal fluctuation (e.g., those than can happen under daily use and not necessarily extreme conditions such as a power blackout), between the time of taking the first satellite image (before the outage) versus the time of taking the second satellite image (during the outage). In that case, an area of interest is deemed to have experienced *outage* if black pixel ratio is greater than 1.0. However, due to the presence of normal fluctuations in loads, there can be areas with a black pixel ratio greater than 1.0 that have not experienced outage; they have rather only experienced less loading, unrelated to the outage. Accordingly, sensitivity to such unrelated factors was reduced by examining four different cut-off points: 1.0, 1.1, 1.2, and 1.5. For the purpose of this study, the 1.2 cutoff point was chosen for two reasons: first it displayed the highest number of significant variables based on t-test results compared to the rest which can be a good indication of its higher accuracy in depicting actual outages; and second, random manual visual inspection of various raster files exhibited more accurate depiction of the actual outage under this setting.

Then the effect size of the difference for the variables that had statistically significant different means between neighborhoods with power outages compared to those with power during Winter Storm Uri were determined (See Table 2).

*Table 2: T-test results for various Cut-Points*

| | 1.5 Cut Point – T-test results – statistical significance | 1.2 Cut Point – T-test results – statistical significance | Cohen's D for the 1.2 Cut Point t-tests | 1.1 Cut Point – T-test results – statistical significance | 1.0 Cut Point – T-test results – statistical significance |
|---|---|---|---|---|---|
| *Age: 65 and over,%* | *** | *** | -.42 | *** | * |
| *Race: Black or African American, %* | NS | NS | | NS | NS |
| *Race: Asian, %* | NS | * | -.18 | * | NS |
| *Hispanic %* | NS | * | .17 | NS | NS |
| *Vacant Housing, %* | *** | *** | .37 | *** | NS |
| *1 unit, Housing Structures, %* | *** | *** | **-1.0** | *** | * |
| *20 or more units, Housing Structures, %* | *** | *** | **+1.0** | *** | * |
| *Housing Built before 1979, %* | NS | NS | | NS | NS |
| *Owner Occupied Housing, %* | *** | *** | **-.89** | *** | NS |
| *No Vehicle Access (owner-occupied housing), %* | *** | *** | **.56** | *** | NS |
| *Over Crowded Housing, %* | *** | ** | .32 | ** | NS |
| *In Labor Force, %* | *** | *** | .47 | *** | NS |
| *Unemployment Rate* | NS | ** | -.27 | * | NS |
| *Public Transit for Work Commute, %* | *** | *** | **.53** | *** | NS |
| *Retail Employment, %* | * | * | -.20 | NS | NS |
| *FIRE Employment, %* | NS | * | .20 | NS | NS |
| *Median Household Income, $* | *** | * | -.22 | * | * |
| *No Health Insurance, %* | *** | ** | .34 | * | NS |
| *Poverty Level, all* | *** | *** | .40 | *** | NS |
| *Poverty, age 65 and over, %* | * | ** | .37 | *** | NS |
| *Household Size* | *** | *** | **-.61** | *** | ** |
| *Family Size* | *** | *** | -.42 | *** | * |
| *Education: High School Grad or equivalent, %* | * | ** | -.24 | ** | * |
| *Education: Bachelor's degree, %* | NS | NS | .16 | NS | * |
| *Mobility: Lived in Same House for at least 1year, %* | *** | *** | **-.57** | *** | NS |
| *Linguistically Isolated (speak English less than "well"), %* | *** | *** | .46 | *** | NS |

According to the test results for variables with large effect size (0.5 and above) it can be inferred that neighborhoods (census tracts) with a power outage had *fewer single-family housing*, *more multi-family housing*, *fewer owner-occupied housing*, *household size* and *more users of public transit for commuting to work* together *with more newcomers* than neighborhoods without a power outage. Also, among variables with medium effect size (0.3-0.5), results indicated that neighborhoods with a power outage had *more linguistically isolated people, more people living in poverty, more people with no health insurance, more people in the labor force, more vacant housing, more overcrowding, smaller family size, and fewer seniors* than neighborhoods without a power outage.

*Logistic Regression*

Identification of parameters with large and medium effect size such as *housing type, public transit work commute, mobility*, etc. led to the next part of the study investigation to explore the significance of each of these parameters in predicting the category of each census tract with regards to their level of impact/outage. More specifically, this was carried out through the application of binary logistic regression in which changes in the log odds of belonging to an affected census tract per a unit increase in predictor variables was investigated. Odds refer to the probability ratio of being in affected group versus unaffected group. Several binary logistic regression models were run using stepwise conditional forward selection testing in SPSS v.27. Through this method, parameter entry is tested based on the significance of score statistic while removal testing is based on the probability of a likelihood-ratio statistic founded on conditional parameter estimates (SPSS,

2021). Among the developed models, a model with four variables as shown in Table 3 was selected after accounting for simplicity, interpretability, and goodness of fit. Included parameters consisted of *percentages of one-unit structures*, *public transportation users*, *linguistically isolated people*, *and high school graduates* within census tracts. The model had a Nagelkerke pseudo–R Square of 0.261 which was marginally lower than the model with the inclusion of all the variables (in percentages as shown in Table 1) while sharing the same significant variables. Additionally, calculated Chi-square for Hosmer and Lemeshow test turned out to be insignificant (Chi-square 7.02, p-value 0.535), indicating the model's goodness of fit (Hosmer Jr, Lemeshow, & Sturdivant, 2013). It is worth noting that having a relatively low R square is the norm for logistic regression; however, they are being suggested to be used as a statistic to compare and evaluate various competing models (Hosmer Jr et al., 2013). Results from the logistic regression indicated the positive impact of *linguistically isolated people* and *public transport commute percentages* in increasing the log odds of belonging to affected census tracts. As percent of *linguistically isolated population* per census tract turns out to have a highly significant correlation with the percent of Hispanic population within the same census tract (Pearson correlation of 0.824 at 0.01 level) this can be an indication of disproportionate impact among various ethnicities within the county. The same positive impact can be seen in the percentage of population taking public transport to work. On the other hand, higher concentration of *one-unit structures* and, subsequently, less concentration of *multifamilyhousing*, appeared to have a negative impact implying that an increase in the percentage would increase the log odds of belonging to the unaffected category. The same applies to percentage of *high school graduates* within a census tract, which resulted in a negative impact. These impacts are visualized in Figure 5 to show how probability of belonging to an affected census tract would be affected by changes in any of these parameters when holding the rest of the parameters unchanged at their mean. As shown in Figure 5, an increase in the percentage of public transportation users for work commute together with the percentage of linguistically isolated population within a census tract increase its probability to be located among the affected census tracts. On the other hand, this probability decreases as the percentage of single-family residences and high school graduates increases within a census tract.

*Table 3. Variables in the equation*

|  | B | S.E. | Wald | Sig. | Exp(B) |
|---|---|---|---|---|---|
| *OneUnitStructure%* | -0.027 | 0.004 | 47.652 | <.001 | 1.028 |
| *PublicTransport%* | 0.067 | 0.032 | 4.455 | .035 | .935 |
| *LinguasticallyIsolated%* | 0.029 | 0.007 | 18.745 | <.001 | .971 |
| *HSgrad%* | -0.048 | 0.013 | 14.196 | <.001 | 1.049 |
| *Constant* | -0.087 | 0.322 | 0.073 | <.001 | 1.091 |

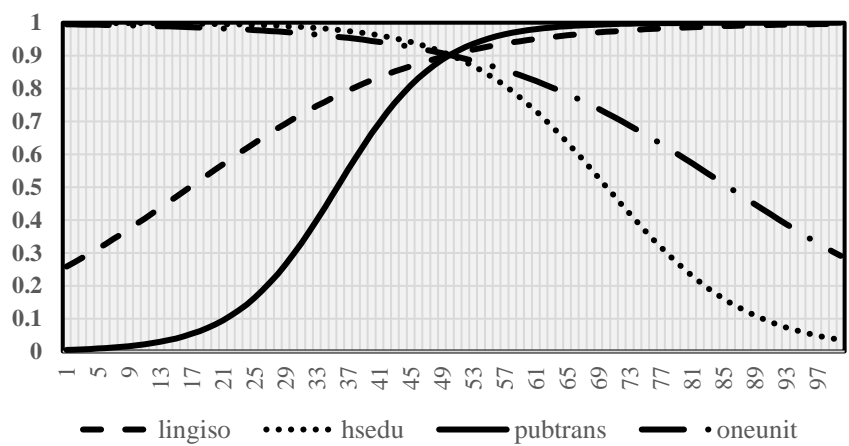

*Figure 5. Probability of belonging to affected category – Sensitivity analysis*

*Principal Component Analysis*

Finally, principal component analysis (PCA) was performed using SPSS v.27 to reveal how percentage of black pixels loads on various components within our data sets. Similar to logistic regression, all the variables in percentages were used in this analysis. Number of factors was limited to four (see Figure 6 and Table 3) as they cumulatively cover at least 60% of the variance and factors beyond four encompass less than five percent of the variance (Bartholomew, Steele, & Moustaki, 2008; Hair, 2009; Peterson, 2000). As shown in Table 4, results from PCA revealed noticeable positive loading of black pixel ratio on a factor which has been additionally loaded by percentage of African American population, public transport commute, poverty, vacant housing, multifamily housing, and no vehicle owner occupied units, a result which is impartially aligned with the previous approaches. Even though these results are not conclusive, they suggest possible presence of disproportionate impacts on low-income communities of color within Harris County.

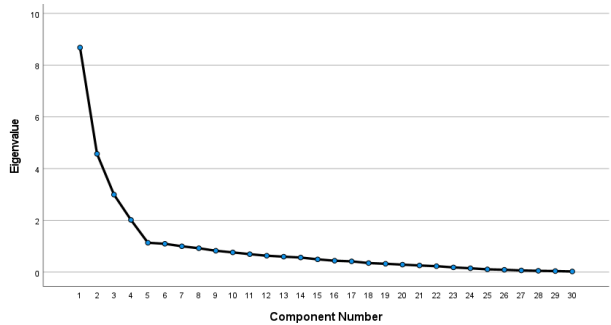

*Figure 6.* Scree plot

*Table 3.* Variables in the equation

| Component | Initial Eigenvalues | | |
|---|---|---|---|
| | Total | % of Variance | Cumulative % |
| 1 | 8.680 | 28.933 | 28.933 |
| 2 | 4.572 | 15.240 | 44.173 |
| 3 | 2.996 | 9.986 | 54.159 |
| 4 | 2.017 | 6.725 | 60.884 |
| 5 | 1.134 | 3.780 | 64.664 |
| 6 | 1.095 | 3.650 | 68.314 |

*Table 4.* Variables in the equation

| | Component | | | |
|---|---|---|---|---|
| | 1 | 2 | 3 | 4 |
| *Age_85andover_PCT* | -.164 | .158 | -.258 | .665 |
| *Age_65andover_PCT* | -.328 | -.170 | -.154 | .837 |
| *Race_BlackAfricanAmericanr_PCT* | -.152 | .245 | .808 | .161 |
| *Race_Asian_PCT* | -.368 | .178 | -.291 | -.106 |
| *Hispanic_PCT* | .908 | -.142 | -.026 | -.203 |
| *VacantHousingUnits_PCT* | .101 | .594 | .069 | .200 |
| *1_UnitStructure_PCT* | -.122 | -.878 | .036 | .256 |
| *20ormoreunitsStructure_PCT* | -.095 | .821 | -.243 | -.142 |
| *Builtbefore1979_PCT* | .598 | .052 | .045 | .447 |
| *OwnerOccupied_PCT* | -.294 | -.853 | -.184 | .215 |
| *NoVehicles_OwnerOcc_PCT* | .349 | .592 | .424 | .298 |
| *LackPlumbing_OwnerOcc_PCT* | .122 | .103 | .120 | .147 |
| *OverCrowded_1.5_PCT* | .499 | .150 | -.100 | -.049 |
| *InLaborForce_PCT* | -.318 | .250 | -.228 | -.707 |
| *Unemployed_PCT* | .074 | .025 | .640 | .004 |
| *PublicTrans_Commute__PCT* | .076 | .565 | .333 | .160 |
| *Retail_Employment__PCT* | .127 | -.100 | .460 | -.202 |
| *FIRE_Employment__PCT* | -.515 | .098 | -.404 | .096 |
| *NoHealthInsurance__PCT* | .867 | .139 | .197 | -.182 |
| *Poverty_Family__PCT* | .729 | .373 | .405 | .061 |
| *Poverty_FemaleHHwithkids__PCT* | .667 | .175 | .273 | .062 |
| *PovertyLevel_all_PCT* | .728 | .408 | .400 | .084 |
| *PovertyLeve_65andover_PCT* | .531 | .297 | .207 | .199 |
| *FemaleHHwithkids_PCT* | .357 | .108 | .573 | -.091 |
| *Households_65andover_PCT* | -.127 | -.409 | .046 | .827 |
| *Education_HighSchoolGrad_PCT* | .564 | -.190 | .626 | -.006 |
| *Education_BachDegree_PCT* | -.765 | .146 | -.498 | .012 |
| *Mobility_SameHouse1year_PCT* | .278 | -.704 | .030 | .223 |
| *LinguisticallyIsolated_PCT* | .879 | .067 | -.092 | -.191 |
| **BPP_ratio** | **.131** | **.253** | **-.123** | **-.092** |

Rotation Method: Varimax with Kaiser Normalization, Rotation converged in 8 iterations

## CONCLUSIONS AND FUTURE WORK

While the results from this study suggest disproportionate impacts among populations with various demographic and socioeconomic statuses, the question remains on how the engineering community, utilities, and policymakers can address inequities and ultimately enhance resiliency in areas that are proven to be affected disproportionately during Winter Storm Uri and other extreme events. In response, two potential power engineering interventions are proposed.

First, as it was observed in this study, when looking at the differences between neighborhoods, it was found that the neighborhoods that had power outage were disproportionally vulnerable. They had more multifamily housing, overcrowded housing, lower owner occupancy, more persons with limited English speaking, more persons without access to a car, more persons who rely on public transit for work commuting, and more persons who recently moved into the neighborhood. These results are particularly insightful within the recovery period. That is, even though different areas were affected similarly when the disaster occurred, the areas with lower household income and higher percentage of ethnic minorities remained without power for a longer period while higher-income predominantly non-Hispanic White areas recovered more quickly. This could be due to various factors, such as lack of more advanced technologies such as Fault Location, Isolation, and Service Restoration (FLISR) (DOE, 2014), which can significantly accelerate service restoration due to an automated ability to pinpoint the points of failure in order to assist utility personnel to restore service faster. More investment in vulnerable geographic areas might be needed to help mitigate disparities in grid resiliency.

Second, there are evolving technologies that can help maintain electricity service during extreme events in critical community resources such as at hospitals, shelters, schools, churches, etc. In particular, recent advancements in the area of microgrid technologies, in combination with the installation of onsite renewable generation and energy storage resources, are improving the ability to sustain *isolated* operation of a critical facility for several days, thereby serving the affected community until service is fully restored, e.g., see (Li, Shahidehpour, Aminifar, Alabdulwahab, & Al-Turki, 2017; X. Liu et al., 2016).

As part of our future work, the authors plan to investigate how critical community resources were affected during Winter Storm Uri and similar extreme events to gain a clearer understanding of disparities in resilience related to specific types of critical infrastructure. The results will help further identify the engineering challenges and potential solutions required to eliminate existing demographic disparities associated with the response to and restoration of disaster-caused electrical outages.

As extreme weather events like Winter Storm Uri become more frequent, intense, and unpredictable due to climate change, it is important to understand how they impact critical infrastructures like the power grid and how such impacts are compounded by socioeconomic and racial inequalities. Our analysis of spatiotemporal and demographic data found that geographic areas with a higher percentage of single family homes recovered from the power outages that occurred during Winter Storm Uri and possessed lower rates of remaining outages during the latter stages of the recovery/restoration phase than areas with a higher proportion multifamily housing communities. Understanding these disparate impacts of Winter Storm Uri is integral to developing appropriate response, recovery, and mitigation plans for future events that disrupt the power grid. Our findings could assist utilities and government entities to enact more equitable approaches to managed service outages and power grid resiliency in the face of natural hazards. It is worth noting that this study is not immune to limitations. More specifically, even though results suggested the importance of community characteristics in how they were affected by the outage they are not conclusive due to lack of high-resolution longitudinal outage data, limited public data on the grid conditions during the storm, etc.; thus, requiring a follow up confirmatory study that can collect data throughout the outage.

**ACKNOWLEDGEMENTS**

This research was supported in by the National Science Foundation awards CMMI 2141203, 2141092, 2143902, for which the authors express their appreciation. Publication of this paper does not necessarily indicate acceptance by the funding entities of its contents, either inferred or explicitly expressed herein.